\documentclass[compsoc]{IEEEtran}
\pdfoutput=1
\usepackage{cite}
\usepackage{amsmath,amssymb,amsfonts}
\usepackage{graphicx}
\usepackage{textcomp}
\usepackage{xcolor}
\usepackage{hyperref}

\usepackage{caption}
\usepackage{subcaption}
\usepackage{multirow}
\usepackage{booktabs}
\usepackage{fancyhdr}
\usepackage[vlined, ruled, linesnumbered]{algorithm2e}
\usepackage{algpseudocode}
\SetCommentSty{mycommfont}
\SetKwInOut{Input}{input}
\SetKwInOut{Output}{output}
\usepackage{cleveref} 

 \crefname{section}{Section}{Sections}
 \crefname{theorem}{Theorem}{Theorems}
 \crefname{lemma}{Lemma}{Lemmas}
 \crefname{equation}{Equation}{Equations}
 \crefname{proposition}{Proposition}{Propositions}
 \crefname{claim}{Claim}{Claims}
\crefname{appendix}{Appendix}{Appendices}
   \crefname{algorithm}{Algorithm}{Algorithms}
 \crefname{figure}{Figure}{Figures}
 \crefname{table}{Table}{Tables}
 \crefname{remark}{Remark}{Remarks}
 \crefname{definition}{Definition}{Definitions}
 \crefname{equatinon}{Equation}{Equations}
 \crefname{corollary}{Corollary}{Corollaries}
\usepackage{amsthm}
\usepackage{lipsum} 

\usepackage[sort,comma,numbers]{natbib}

\def\BibTeX{{\rm B\kern-.05em{\sc i\kern-.025em b}\kern-.08em
    T\kern-.1667em\lower.7ex\hbox{E}\kern-.125emX}}

\author{\IEEEauthorblockN{Mingxue Xu\IEEEauthorrefmark{1},
Tongtong Xu\IEEEauthorrefmark{2}, Po-Yu Chen\IEEEauthorrefmark{1}}\\
\IEEEauthorblockA{\IEEEauthorrefmark{1} Imperial College London, London, UK\\
\IEEEauthorrefmark{2} Baidu Inc., Beijing, China}

Email: \IEEEauthorrefmark{1}{\{m.xu21, po-yu.chen11\}}@imperial.ac.uk,
\IEEEauthorrefmark{2}xutongtong@baidu.com}

\begin{document}

\title{Private Training Set Inspection in MLaaS}

\newtheorem{example}{Example}
\newtheorem{theorem}{Theorem}
\newtheorem{definition}{Definition}
\newtheorem{assumption}{Assumption}

\maketitle

\begin{abstract}
Machine Learning as a Service (MLaaS) is a popular cloud-based solution for customers who aim to use an ML model but lack training data, computation resources, or expertise in ML. 
In this case, the training datasets are typically a private possession of the ML or data companies and are inaccessible to the customers, but the customers still need an approach to confirm that the training datasets meet their expectations and fulfil regulatory measures like fairness.
However, no existing work addresses the above customers' concerns. 
This work is the first attempt to solve this problem, taking data origin as an entry point.
We first define origin membership measurement and based on this, we then define {\bf diversity} and {\bf fairness} metrics to address customers' concerns. 
We then propose a strategy to estimate the values of these two metrics in the inaccessible training dataset, combining shadow training techniques from membership inference and an efficient featurization scheme in multiple instance learning. 
The evaluation contains an application of text review polarity classification applications based on the language BERT model.  
Experimental results show that our solution can achieve up to 0.87 accuracy for membership inspection and up to 99.3\% confidence in inspecting diversity and fairness distribution.
\end{abstract}


\maketitle

\section{Introduction}\label{sec:intro}

% 0. Background
Over the past decade, Machine Learning (ML) has envolutionised many fields and formed an ML market valued at USD 16.2 billion in 2021~\cite{MachineL66:online}. This trend prompts subareas such as Machine Learning as a Service (MLaaS).
MLaaS is an outsourced service that contains model training, sometimes also involves building datasets that meet customer requirements. Building such datasets often requires considerable manpower and resources, thus in many cases the MLaaS provider only releases the model access to the customers but no access to the training dataset~\footnote{\url{ https://azure.microsoft.com/en-us/services/cognitive-services}}\footnote{\url{ https://www.ibm.com/cloud/watson}}\footnote{\url{https://cloud.google.com/solutions/marketing-analytics}}. 
This kind of service is suitable for customers with limited budgets but still want to personalize the training dataset. 

% 1. Problems
On the customer side, there are two issues of profit. The first issue is {\bf the dishonest claim of the training data properties from the MLaaS provider}.
In other words, the MLaaS provider might lie that they have the required data from the customer but, in fact not. 
This dishonesty is difficult to detect without direct access to the training set. The performance of the ML model in the actual deployed environment is potential evidence but, since the data samples in the actual deployed environment do not overlap with the original training set, good ML model performance is not a sufficient condition for an honest claim by the MLaaS provider.
A model trained on a public dataset may overall outperform that on the extra-charged private training set in the actual deployment environment. The second issue is {\bf a low-cost measurement in terms of legal requirements (i.e. fairness)}. On the MLaaS provider side, the collection of fair data costs extra manpower and resources. However, monitoring either data collection or model production is a daunting task. And similar to the dishonest claim by the MLaaS provider, the unfairness of the training dataset is also difficult to detect through the model prediction in the actual deployed environment.

% 2. Solution Summary
To address these two issues, we choose a critical aspect of data management as an entry point - {\bf data origin}, which indicates the entities related to {\it data generation} (e.g. movies that the reviews describe). 
Here we introduce a process named {\it inspection}, to check the above customers' concerns through the measurements of data diversity and fairness, with data the origin existence in the training set (i.e. membership) as a bridge. 
Data diversity, and also data origin membership is related to the data properties claim by the MLaaS provider, since it costs extra for the data collection of specific data origins and a wide range of data origins.
Data origin fairness is related to regulations like biased/unbiased decision-making, like the data proportion of the protected group in the overall population. 
This proposed inspection is made by a third party named {\it inspector}, an intermediary between the customer and service provider, on behalf of the customers' interest. Section~\ref{sec:roles} describes the roles and capabilities of the inspector, customer and service provider in real-world settings. 

% 3. Recent Research
Some existing research addresses similar problems, yet none exactly states these two issues. There exists a personalized data sources selection strategy for the customersr~\cite{tailor_provenance}, but the selected datasets can finally be fully accessed by the customers. On the other hand, in our setting, the customer, as well as the inspector, have no access to the private training set. Other work enabled user-level Membership Inferences (MI) via only access to ML models~\cite{Song2019AuditingDP, Miao2021TheAA, jaiswal2020privacy}, but did not mention data diversity and fairness. This work aims to fill this gap.

\begin{figure*}
    
    \centering
    \begin{subfigure}{0.33\linewidth}
         \centering
         \includegraphics[width=\linewidth]
         {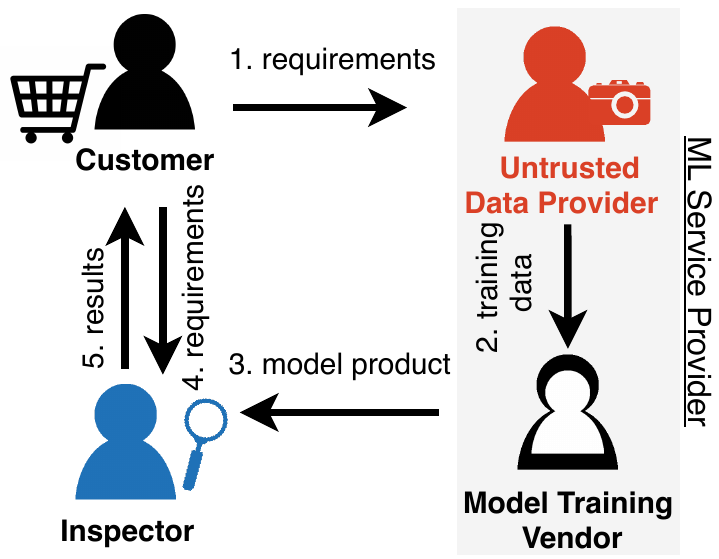}
         \subcaption{Inspection Protocol}
         \label{fig:settings}
     \end{subfigure}
     \begin{subfigure}{0.33\linewidth}
         \centering
         \includegraphics[width=\linewidth]{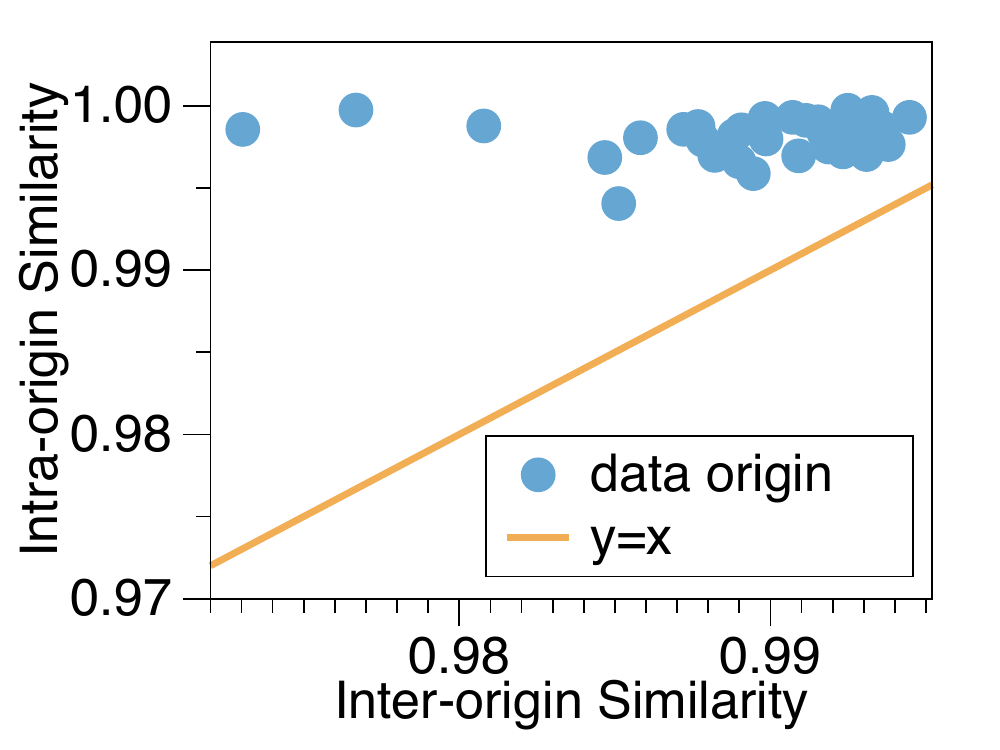}
         \subcaption{Data Similarities}
         \label{fig:insight}
     \end{subfigure}
     \hfill
     \begin{subfigure}{0.32\linewidth}
         \centering
         \includegraphics[width=\linewidth]{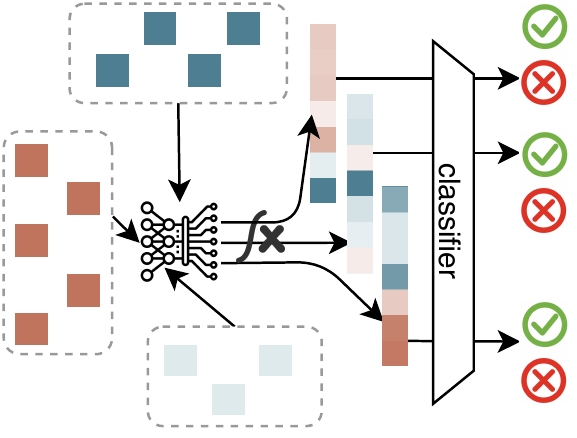}
         \subcaption{Origin-level Inference}
         \label{fig:pipe}
     \end{subfigure}

    \caption{Settings and empirical assumption of the inspection. (a) Inspection in ML services. The model manufacturer might unitely or separately contain data provider\protect\footnote{\url{https://ieee-dataport.org/}} and model training vender\protect\footnote{\url{https://cloud.google.com/automl}}. In this work, we assume the model training vendor, the inspector and the customers are all benign, while the data provider is untrusted. (b) Demonstration that samples from the same origin have larger cosine similarity than those from different origins, indicates that data are distinguishable related to their origin. (c) The difference between classic sample-level Membership Inference and origin-level Membership Inference, where samples are grouped according to their data origins.}\label{fig:main-idea}
\end{figure*}

% 4. Methodologies
In this work, we implement inspection with defined practical metrics. To this end, we first build up three metrics to describe dataset properties on the data origin level, in terms of membership, diversity and fairness. We then extend the original shadow learning algorithm from sample-level MI~\cite{MIA} to the data origin level, to obtain the origin membership in the training set.
We assume that the same origin data keep a detectable and learnable invariant pattern compared with other origins, which is empirically validated in Figure~\ref{fig:insight}.
Based on this assumption, we group data samples according to data origin. 
These grouped samples are input to the ML model together, and then converted into set-level features, ultimately input to a set-level MI binary classifier, as shown in Figure~\ref{fig:pipe}. 
With the origin membership inspected, we obtain the estimated value of the metrics we proposed. Finally, the proposed framework is evaluated over a text review application.

Our main contributions are summarized as follows:
\begin{itemize}
    \item To our best knowledge, we are the first to investigate {\it private} training set diversity and fairness in MLaaS. We formally define the problem and propose an inspection framework, taking data origin as an entry point (Section~\ref{sec:prob}).

    \item Assuming that the inspector can randomly sample the testing data origins and based on our previous work~\cite{xu2022data}, we propose a strategy to investigate dataset diversity and fairness, which combines shadow training and multiple instance learning  (Section~\ref{sec:methodology},~\ref{sec:implementation}).
    
    \item We take movie review polarity classification as a case study. The concerned data origin is the movie, and the model product is a widely exploited DNN model - Small BERT~\cite{turc2019}. The experiment evaluation shows that our proposed methodologies achieve significant performance, with 99.3\% confidence in inspecting the distribution of data diversity and fairness distribution on the data origin level (Section~\ref{sec:exp}, \ref{sec:eval}).
\end{itemize}
\section{Model and Problem Statement}\label{sec:prob}
The work aims to inspect private training sets diversity and fairness in MLaaS. Focusing on this problem, this section clarifies the following points: 
1) the roles and their corresponding capabilities in real-world MLaaS inspection, and the informal definition;
2) the exact definition of training sets diversity and fairness, and the reason for choosing data origin as an entry point to inspect them, and therefore, what are the sub-objective functions; 
3) based on data origin, what metrics are we measuring for the inspection?  

\subsection{Roles and Capabilities in Real-world Settings}\label{sec:roles}

Figure~\ref{fig:settings} gives the overview of the MLaaS with the inspection. In the following we will introduce the model of the inspection process and the involved roles' capabilities, and their instances in real-world applications. For clarity, we split the ML service provider into two roles: data provider and model training vendor. The model training vendor is trusted and honestly follows the protocol as promised to the customer. Herein, we discuss these four roles:
\begin{enumerate}
% actual instnaces
    \item {\bf Customers}: Individuals or small and medium enterprises (SMEs) who might know their business well, yet do not have enough data and ability (e.g. computation resources or ML expertise) to train the ML models. 
    \item {\bf Inspector}: The individuals/SMEs with more data and computation resources compared with the customer, larger enterprises with certain expertise of the customer’s business, or professional institutions authorized by the government. The inspector has the same model access as the customers. % The difference among these three instances lies in the inspect targets, which will be further explained in xxx.
    \item {\bf Model Training Vendor}: Enterprises with the computation resources and ML expertise to support ML model training.
    \item {\bf Data Provider (Untrusted)}: Enterprises that take the responsibility to collect data required by the customer, yet may lie to the customer about whether they hold the required data.
\end{enumerate}

The inspectors of different scales have different capabilities (data and computation resources), and take responsibility for the different inspection metrics.
Before going into the details of these settings, we shall first propose the finer definition of the core concept in this work - data origin, and the description of model production.

\subsection{Machine Learning Model Production}

As illustrated in Figure~\ref{fig:settings}, the data provider provides the target training set ${D}^\text{target}$. The model training vendor then trains an ML model $f(\cdot)$ with their own computation resources.
With this trained model $f(\cdot)$ (as well as the hyperparameters) and an arbitrary input instance set $X$, the customer (also the inspector) would have full access to $f$ and the intermediate outputs of all the layers of $f$, that is
\begin{equation}
f^{(i)}(X),\quad i=0,1,\ldots,n-1.
\end{equation}
After being inspected by the third-party inspector $\mathcal{I}$, this model $f(\cdot)$ will then be sold on the MLaaS market or delivered to the customer.

\subsection{Diversity and Fairness Metrics of a Dataset}\label{subsec:diversity-fairness}

As explained in Section~\ref{sec:intro}, both diversity and fairness are related to data collection costs. This section starts from the diversity and fairness measurement definition on the whole training set level, and then uses the measurement of available data origins to approximate. 

{\it Diversity} here describes the extent of relative data samples even distributed in the plausible sample space. Since this metric appear along with the fairness metric in this work, we should note that it is not the same as ``diversity'' in demographic (e.g. race and age), but share similar statistics characters like data sample variance. A straightforward diversity measurement would be through calculating the mean signed deviation of the scalarized samples in the dataset. ``Scalarized'' means using a scalar to represent each data sample, which is too coarse-grained. On the other hand, this method requires the mean value to be applied in every term of the polynomial expression, thus it is very sensitive to the occurrence of outliers. Here, we average the cosine similarity of any two data samples for finer-grained and enhanced robustness.
\begin{definition}[Dataset Diversity Metric]\label{def:set-diversity} The sample diversity metric $\bar d$ of a data input set $X$ is given by 
\begin{equation}
    \bar d = \frac{1}{|X|}\sum^{|X|}_{i=1}\frac{1}{|X-1|}\sum^{|X|}_{j=1, j \neq i}\texttt{CosSim}({\bf x}^{(i)}, {\bf x}^{(j)})
\end{equation}
  where ${\bf x}^{(i)}$, ${\bf x}^{(j)}$ are the $i$th and $j$th flattened data input samples, respectively. %$\bf{\bar x}$ is the mean vector of all the flattened data input samples in $X$, where $\texttt{CosSim}(\bf{x}^{(i)}, \bf{\bar x}) = \frac{{{\bf x}^{(i)}}\cdot {\bf{\bar x}}}{|\bf{x}^{(i)}||\bf{\bar x}|}$ is the cosine similarity.
  $\texttt{CosSim}({\bf x}^{(i)}, {\bf x}^{(j)}) = \frac{{{\bf x}^{(i)}}\cdot {{\bf x}^{(j)}}}{|{\bf x}^{(i)}||{\bf x}^{(j)}|}$ is the cosine similarity.
\end{definition}

{\it Fairness} considered in this work refers to group fairness~\cite{Zemel2013LearningFR}, meaning each group that shares the same attribute (e.g. gender or race) should be treated similarly in the ML model prediction. Therefore literally, we have the following dataset fairness measurement.

\begin{definition}[Dataset Fairness Metric]\label{def:set-fairness}
Considering a sensitive attribute $A=\{a, \bar a\}$, the overall fairness metric $b$ of a given dataset, $D=\{(x,y)\}$, is given by
\begin{equation}
b=\frac{\sum_{(x,y) \in D^{a}} y}{|D^{a}|} - \frac{\sum_{(x,y) \in D^{\bar a}} y}{|D^{\bar a}|}
\end{equation}
where $D^{a}=\{(x,y)| (x,y) \in D, A(x)=a \}$ and $D^{\bar a}=D-D^{a}$. 
\end{definition}

In Definision~\ref{def:set-fairness}, we do not use the absolute value form, since both the positivity and negativity of $b$ convey rich information about the bias. For instance, if we set $D^{a}$ as the dataset of the protected group, $b$ near $-1$ indicates that $D$ is an unfair dataset for the protected group, while $b$ near $1$ implies the $D$ is an unfair dataset for the unprotected group. 

\section{Inspection on Data Origin Level}\label{sec:methodology}

The overall objective of this work is to inspect the diversity and fairness of unaccessible training datasets. In a realistic setting, there are no exact data samples for the inspection, thus we take data origin as an entry point. We first give the definition of data origin, and then build up the sub-objective metrics (data origin level) that can be used to approximate dataset-level diversity and fairness metrics defined in Definition~\ref{def:set-diversity} and~\ref{def:set-fairness}. For these data origin-level metrics, we give the required assumptions that support the data origin level to facilitate the dataset-level inspection, and the real-world instances of the customer and the inspector.  

\subsection{From the Data Origin to the whole Dataset}\label{sec:gt-def}

The literal definition of data origin is given in Section~\ref{sec:intro}, or in other words, ``where the data is
generated or what subject the data describe''. For the mathematical definition of data origin, please refer to our previous work~\citep[Definition 1]{xu2022data}. In this section, we clarify the assumptions of using data origin level measurement to estimate the diversity and fairness of the whole dataset, and afterwards data origin level metrics that facilitate the dataset inspection. The root assumption is:
\begin{assumption}\label{assumption:root}
The inspector can randomly sample the data origin on the whole dataset population; thus, the sampled data origins represent the training set.
\end{assumption}
Before proceeding further, we can have the degraded data diversity and fairness metrics on the data origin level. However, before this, a data origin level version of Definition~\ref{def:set-diversity} is defined in Algorithm~\ref{alg:ios}. $\mathcal{X}_{V}=\{X_v|v \in V\}$, which denotes a superset consisting of the datasets of each data origin in $V$.

\begin{algorithm}[t]
\SetKwComment{Comment}{$\triangleright$\ }{}
\SetKwInOut{Input}{Input}
\SetKwInOut{Output}{Output}
\SetKwInOut{Initialize}{Initialize}
\SetKwInOut{mean}{mean}

\Input{${V}$, $\mathcal{X}_{V}$, $v_0$}
\Output{origin-level diversity regarding $v_0$ in $D$}
\BlankLine
\Initialize{${S}_{d}\leftarrow\varnothing$}
\For(\tcp*[f]{\small compute normalised input vectors}){$X_v \in \mathcal{X}_{V}$}
{
    ${S}_{0}\leftarrow\varnothing$
    
    $ {\bf x}_v \leftarrow \texttt{flatten}(\texttt{mean}({X}_v)$)
    
    ${\bf x}_v \leftarrow {\bf x}_v / || {\bf x}_v ||_2$
    \tcp*[f]{\small normalization} 
    
    ${S}_t \leftarrow {S}_t \cup \{{\bf x}_v\}$}
\For{
${\bf x}$ $\in$ ${S}_t$ $\setminus$ $\{{\bf x}_{v_0}\}$}
{${S}_{d} \leftarrow {S}_{d} \cup \{ \texttt{CosSim}({\bf x}_{v_0},{\bf x})\}$
}\label{line:dist}

\Return{$\texttt{mean}({S}_{d})$}
\caption{Inter-origin  Similarity  (\texttt{IOS})}\label{alg:ios}
 \vspace{-2px}
\end{algorithm}

\begin{definition}[Inter-origin Similarity of a Data Origin]\label{def:ios}
Given an origin set $V$ and its corresponding data sample input set $X$, the Inter-origin Similarity of a data origin $v$ is defined as
\begin{equation}
    d_v=\texttt{IOS}(V, \mathcal{X}_{V} ,v)
\end{equation}
where \texttt{IOS} is defined in Algorithm~\ref{alg:ios}.
\end{definition}

\begin{definition}[Dataset diversity Metric on Data Origin level]\label{def:ori-diversity} 
Given an origin set ${V}$ of the data input set $X$, the data origin diversity $H^d_{{V}}$ derived from Definition~\ref{def:ios} is given as
\begin{flalign}
 H^d_{{V}} = diag({[{d}_{1}, {d}_{2}, \ldots, {d}_{|{V}|}]}^{T}) 
\end{flalign}
where $d_i=\texttt{IOS}(V, \mathcal{X}_{V} ,v)$ is defined in Algorithm \ref{alg:ios}, $diag(\cdot)$ means to form a vector with the main diagonal elements of the inputted matrix.
\end{definition} 

The reason why $ H^d_{{V}}$ can used to estimate $\bar d$ in Definition~\ref{def:set-diversity} is similar with quality assurance in both manufacturing and service industries, which is actually statistical sampling. 

\begin{figure}
\centering
    \begin{subfigure}{\linewidth}
    \centering
    \includegraphics[width=0.95\linewidth]{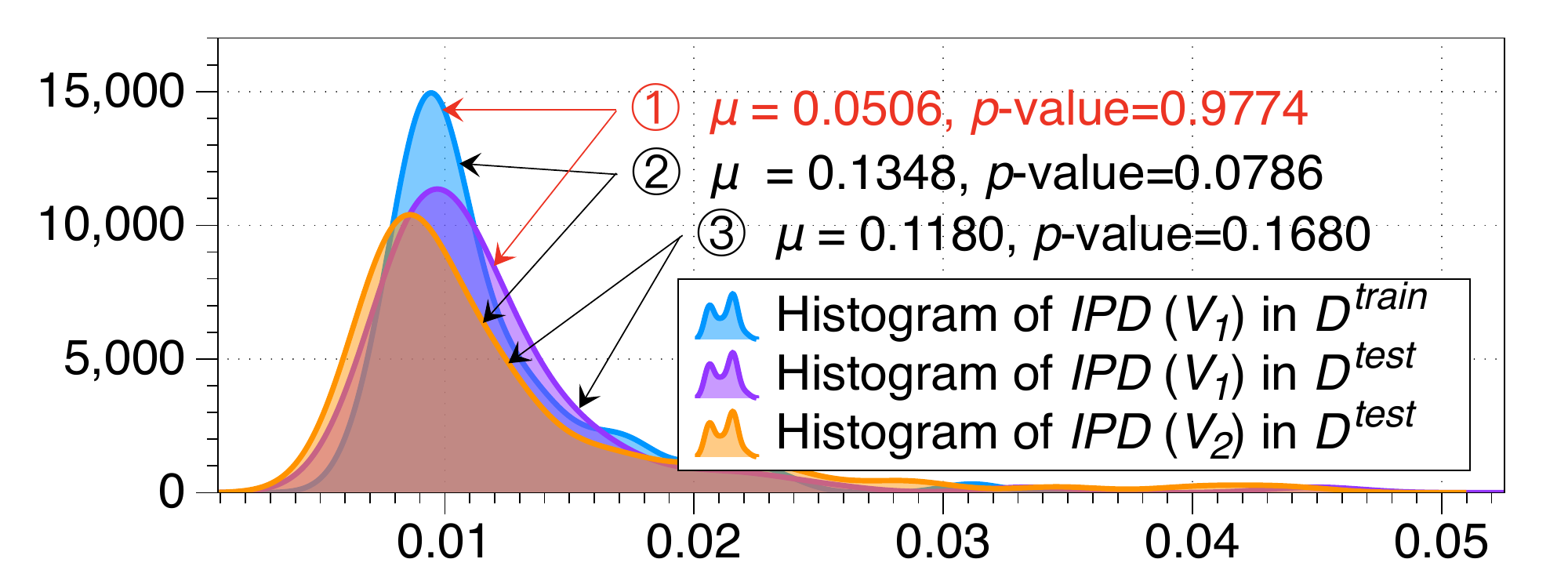}
    \subcaption{Empirical evidence of Assumption~\ref{assumption:maintain}.}
    \label{fig:asm-d-maintain}
    \end{subfigure}

    \begin{subfigure}{\linewidth}
    \centering
    \includegraphics[width=0.95\linewidth]{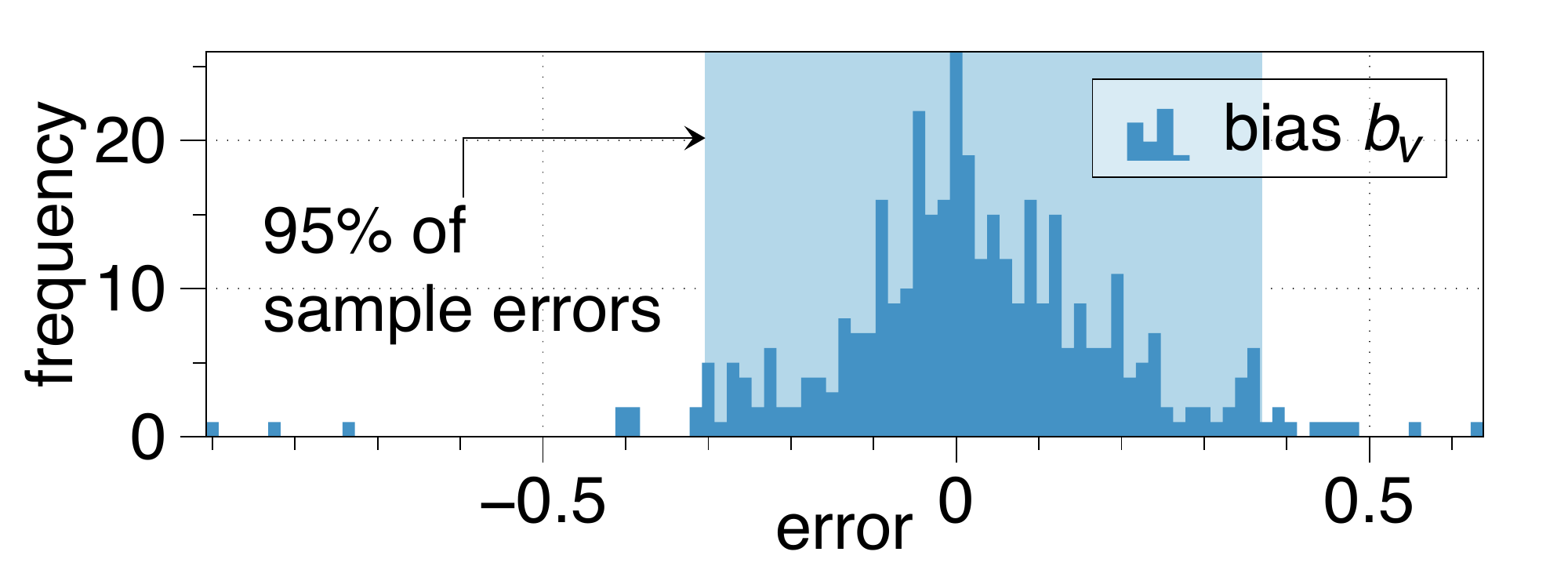}
    \subcaption{Empirical evidence of Assumption~\ref{asm:fair-maintain}.}
    \label{fig:asm-e-maintain}
    \end{subfigure}
\caption{Empirical validations for Assumption~\ref{assumption:maintain}, \ref{asm:fair-maintain}.}\label{fig:assumption}
\end{figure}

\begin{assumption}\label{assumption:maintain}
For an origin, its inter-origin similarity maintains
its distribution across any disjointed dataset, which involves this origin.
\end{assumption}

Assumption~\ref{assumption:maintain} is empirically validated in our preliminary experiment on OpenImage V6 to test if we can approximate inter-origin similarity of ${D}^\text{target}$ with ${D}^\text{test}$ using Kolmogorov-Smirnov test, which is illustrated in Figure~\ref{fig:asm-d-maintain}. 
It should be noted that in this work, we do not consider the factors like sequential sampling that may affect data distribution of origin.

Similarly, we have a degraded fairness metric based on Definition~\ref{def:set-fairness}.
\begin{definition}[Dataset Fairness Metric on Data Origin level] \label{def:ori-behavior-bias}
Given a sensitive attribute $A$ and a set of data samples $D_{v}$ of data origin $v$, $D_{v}$ can be devided into two subsets: $D^{a}_{v}$ and $D^{\bar a}_{v}$. The behavior bias $b_v$ of the origin $v$ is given by
\begin{equation}\label{eq:ori-behave-bias}
b_v=\frac{\sum_{(x,y) \in D^{a}_{v}} y}{|D^{a}_{v}|} - \frac{\sum_{(x,y) \in D^{\bar a}_{v}} y}{|D^{\bar a}_{v}|}.
\end{equation}
Therefore, the dataset fairness metric on the data origin level is
\begin{equation}
 H^e_{{V}} = diag({[{b}_{1}, {b}_{2}, \ldots, {b}_{|{V}|}]}^{T}).
\end{equation}
\end{definition}
As mentioned at the end of Section~\ref{sec:prob}, the higher $|b_v|$ is, the more biased the data origin $v$ behaves. 

Herein, the ultimate goal is to estimate $H^{d}_{V}$ in Definition~\ref{def:ori-diversity} and $H^{e}_{V}$ in Definition~\ref{def:ori-behavior-bias} to approximate $\bar d$ in Definition~\ref{def:set-diversity} and $b$ in Definition~\ref{def:set-fairness}, respectively. 

\subsection{Inspection Framework and Real-world Settings}
This section formally defines the inspection problem based on the possible techniques. In this work, the technical standing point is data origin inference. Additionally, since the inspection framework is application-oriented and requires prerequisites (i.e. the inspector capabilities in the real-world setting) for each component, we give detailed real-world settings along with each component.

\subsubsection{Technical Bedrock: Data Origin Inference}\label{subsec:small}

Data origin inference aims to decide if a data origin is involved in the training dataset, without exact training samples. This technique is used to select possible existing data origins in the training set, when given $H^{d}_{V}$ and $H^{e}_{V}$ in Definition~\ref{def:ori-diversity} and~\ref{def:ori-behavior-bias}, respectively. Adjusted from~\citep[Definition 2]{xu2022data}, we can state the following definition for Data Origin Inference.

\begin{definition} [Data Origin Inference]\label{def:ori}
Data Origin Inference aims to infer the membership of a data origin $v$ in the target training set $D^{\text{target}}$, with an additional data set $D^{\text{aux}}_{v}$ whose origin is $v$, that is

\begin{align}\footnotesize
    \texttt{Ori}(\sim,D^{\text{aux}}_{v},  D^{\text{target}}, \delta,f) = 
    \begin{cases}\footnotesize
    1, &\psi(\sim,D^{\text{aux}}_{v},D^{\text{target}}, f) \geq \delta \\
    0, &\text{otherwise}
    \end{cases}
\end{align}  

\begin{align}\footnotesize
\begin{split}\footnotesize
 \psi(\sim,
  D^{\text{aux}}_{v},D^{\text{target}},f)= 
  \mathbb{P} \left[ \exists D_k \in (D^{\text{aux}}_{v}\cup D^{\text{target}})/\sim, D^{\text{aux}}_{v} \subset D_k \Big|  f(D^{\text{aux}}_{v}) \right ]
\end{split}
\end{align}
$\delta$ is the chosen threshold depending on the application requirements, $\sim$ is the data origin types described in~\citep[Section 3.2]{xu2022data} and $\mathbb{P}$ is the probability.

\end{definition}

Even though Definition~\ref{def:ori} has practical values, there is a case in which the customers know their business well, are also aware of what data origin matters to them, and wonder if the exact origins they care about are involved in the training set. 
They might have a small volume of data samples, of which the attributes (i.e. generation, collection or processing) meet their specific requirements. If customers do not have these data samples, the data samples can be easily obtained on the Internet and used by the inspectors for the inspection.
Small-scale inspectors (individuals/SMEs) are enough for this inspection. They have certain private auxiliary data and enough computation resources to support the general inspection algorithm. However, their auxiliary data are not customised for a certain purpose, thus they cannot directly provide customised ML services.

Consider a practical example for the above case, for a customised shopping reviews' polarity classification for a particular city, the customer may acknowledge that the reviews of certain local stores are essential for their business.
The customer then gives a list of stores to the data provider and also the inspector, the data provider then collect the data, or selects from their existing private datasets of the stores in the list.
If the data is newly collected, the data provider can keep these data for the next time use.
After the ML model product has been delivered, the inspector only needs to check if the stores on the list are involved in the training set.
To facilitate the inspection, the inspector might ask the customer to provide several data samples of these stores, or search the Internet to get these data. 

Because of this practical value of Definition~\ref{def:ori}, we define an inspection type named {\bf Membership Inspection} separately, as follows:

\begin{definition}[Membership Inspection]\label{def:mem-inspect}
Given a target origin set ${V}^\text{tar}$, a test origin set ${V}^\text{test}$, and a set of auxiliary datasets $[D^{\text{aux}}_{1}, D^{\text{aux}}_{2}, \ldots, D^{\text{aux}}_{|{V^\text{test}}|}]$ linked to each $v$ in ${V}^\text{test}$, Membership Inspection is to obtain 

\begin{equation}
H^m_{{V}^\text{test}} = \left[\hat m_{1}, \hat m_{2}, \ldots, {\hat m}_{|{V}^\text{test}|}\right]
\end{equation}
where $m_v=\texttt{Ori}(\sim,D^{\text{aux}}_{v},D^{\text{target}}, \delta, f)$ defined in Definition~\ref{def:ori}.
\end{definition}

\subsubsection{Diversity Inspection}\label{subsec:large}
A straightforward way of diversity inspection is to conceptually refer to Definition~\ref{def:ori-diversity} and practically apply Definition~\ref{def:mem-inspect}.

\begin{definition}[Diversity Inspection]\label{def:diverse-inspect} 
Given ${V}^\text{test}$, ${X}^\text{test}$, Membership Inspection result $H^m_{{V}^\text{test}}$ as defined in Definition~\ref{def:mem-inspect}, Diversity Inspection is to obtain $H^d_{{V}^\text{test}}$ defined as follows
\begin{flalign}
 H^d_{{V}^\text{test}} = diag(  {[{d}_{1}, {d}_{2}, \ldots, {d}_{|{V}^\text{test}|}]}^{T} \times H^m_{{V}^\text{test}}) 
\end{flalign}
where $\times$ denotes cross product, and $d_v=\texttt{IOS}(V^{text}, \{X_v|v\in V^{test}\} ,v)$, where $\texttt{IOS}$ is defined in Algorithm \ref{alg:ios}.
\end{definition} 

To look closer at the practical meaning and motivation of the diversity inspection, take the example of shopping reviews in Section~\ref{subsec:small} once again. If the data provider promises to the customer they would collect data from various city blocks, but in fact, only collects reviews from the same city block, the actual cost would be lower than that of the promise. 
This dishonest behaviour is hard to detect since it is nearly impossible for the inspector to have all the possible data origins in various city blocks.

% naive solution & drawback
A practical way to measure this origin diversity is through data dispersion. Based on this entry point, using the data diversity degree of the data samples is straightforward. If the data samples are dispersed enough, the training data might be from diverse city blocks.
There are popular measurements like statistical variance and standard deviation.
However, such measurements are not suitable for this work, because some origins might be already dispersed in data samples.
For example, an active user of a business review App might give reviews to most businesses (e.g. restaurants, hospitals, sceneries) he/she has been to. Because the reviews cover many business categories, this user's data variance or standard deviation might be higher than a bunch of other users.
In this case, statistical variance and standard deviation are insufficient to address the origin diversity.

Here we use inter-origin data similarity to describe data diversity, and furthermore the data origin diversity. As described in Algorithm~\ref{alg:ios}, for each origin, we use the average cosine distance between the other origin in the dataset to measure to what extent this origin is "different from the others". Suppose the origins in the training dataset are mostly different from each other. In that case, we can conclude the data origin is diverse enough, and the data provider put a relatively high cost to collect diverse data.

In this case, the inspector should be a larger enterprise with certain expertise in the customer's business. They have enough origin data in this field to hold Assumption~\ref{assumption:root}, select out these origins and conduct origin inference. Then they use the predicted member origin inter similarity to summarize the overall origin diversity in the training set. Thus we derive the formal definition of diversity inspection shown in Definition~\ref{def:diverse-inspect}.

\subsubsection{Fairness Inspection}\label{subsec:official}

% Logic of our fairness inspection
The bedrock of fairness inspection is the fairness metric. The fairness metric on dataset and data origin levels have been mentioned in the Definition~\ref{def:set-fairness}, \ref{def:ori-behavior-bias}. Similar to diversity inspection in Definition~\ref{def:diverse-inspect}, we can have the following fairness inspection:
\begin{definition}[Fairness Inspection]\label{def:fairness-inspect}
Given $\mathcal{V}^\text{test}$, $\mathcal{X}^\text{test}$, Membership Inspection result $H^m_{\mathcal{V}^\text{test}}$ as defined in Definition~\ref{def:mem-inspect}, Fairness Inspection is to obtain $H^e_{\mathcal{V}^\text{test}}$ defined as follows:
\begin{equation}
     H^e_{\mathcal{V}^\text{test}} = diag(  {[{b}_{1}, {b}_{2}, \cdots, {b}_{|\mathcal{V}^\text{test}|}]}^{T} \times H^m_{\mathcal{V}^\text{test}}) 
\end{equation}
where $\times$ is cross product, $diag(\cdot)$ means to form a vector with the main diagonal elements of the inputted matrix, and $b_v$ is obtained by Definition~\ref{def:ori-behavior-bias}.
\end{definition}

Let us look closer at Definition~\ref{def:fairness-inspect}. In this work, we consider one typical fairness type - group fairness.
The current group fairness measurement of a dataset is to extract the sensitive features of each data sample and then analyse the correlation between the sensitive features and the prediction results~\cite{Zemel2013LearningFR}. 
% How we do in this work
This idea can be extended at the origin-level, that is to say, the data sets of each origin can be analysed like~\cite{Zemel2013LearningFR} separately.   However, in our ML model production context, the sensitive features of both data samples and origin are not labelled explicitly due to labelling cost concerns.

As described in Assumption~\ref{assumption:maintain} - inter-origin data similarity of origin is likely to be consistent across disjoined datasets, the behaviour bias of origin might also be consistent. 
In the context of fairness, this behaviour bias of origin is when confronting data with sensitive attributes, and the output labels appear differently than that not.
For example, if a data origin has a gender-related preference when giving opinions/predictions, this preference is a kind of behaviour bias.
More specific examples are that a mobile user of a movie review APP tends to give negative reviews when a movie has significant female characters or a mobile user of an image APP like taking photos of females with the label ``kitchenware''. This behaviour bias of data origin has been formally defined in Eq. (\ref{eq:ori-behave-bias}), yet still needs the following assumptions to support Definition~\ref{def:fairness-inspect}, which is valid in Figure~\ref{fig:assumption}.
\begin{assumption}\label{asm:fair-maintain}
For a data origin, its behaviour bias maintains its distribution across any disjointed dataset, involved in this data origin.
\end{assumption}
\begin{assumption}\label{assumption:bias-maintain}
If a data origin has biased behaviour in the collected data, this origin is not suitable for fair datasets.
\end{assumption}

In this case, the inspectors should be professional institutions authorized by the government, which have a large amount of origin and origin data so that the estimation based on sampled origin can be used to estimate the ultimate fairness distribution of the training set.

\section{Implementation}\label{sec:implementation}
\begin{figure*}
    \centering
    \includegraphics[width=1\linewidth]{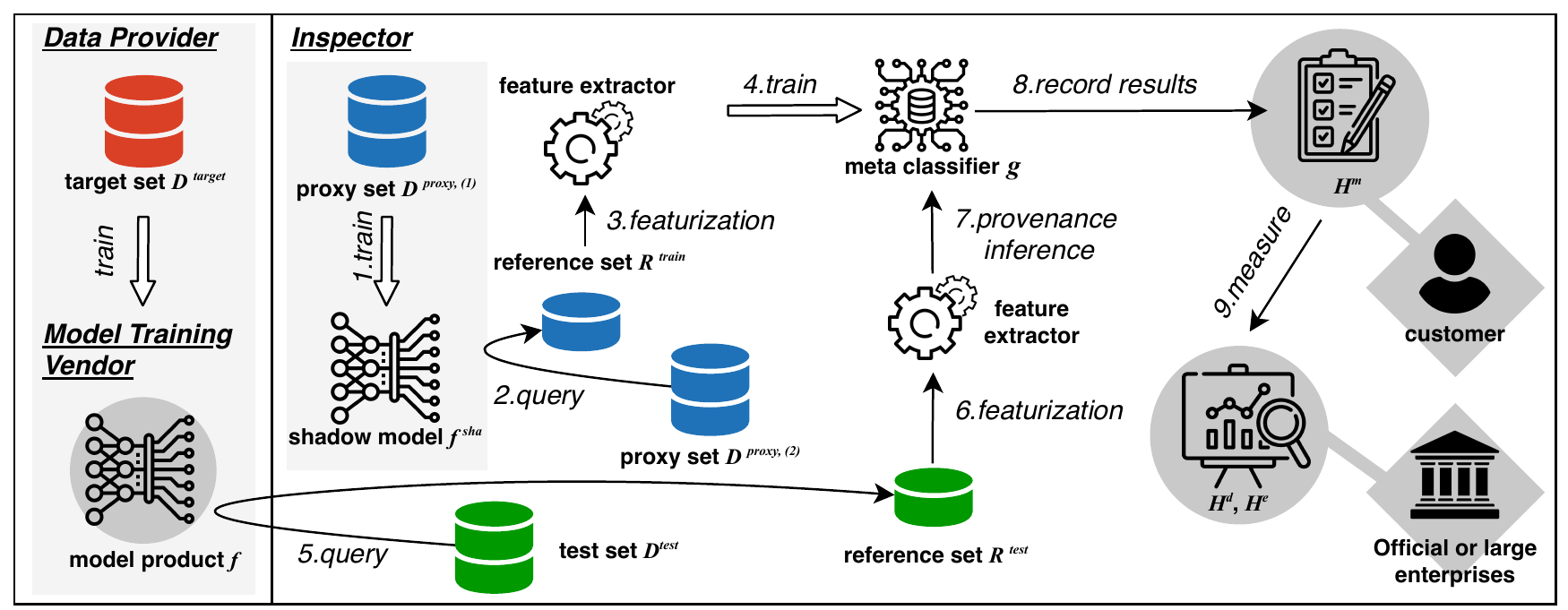}
    \caption{The pipeline of the model production and inspection.}
    \label{fig:pipeline}
\end{figure*}

This section gives the implementation details to enable the inspections proposed by Section~\ref{sec:methodology}.

\subsection{Inspection Pipeline}

Figure~\ref{fig:pipeline} demonstrates the working pipeline of our proposed solution for inspection. 
The inspector first trains a shadow model $f^\text{sha}(\cdot)$ with the same architecture as the target model $f$ with a local proxy dataset ${D}^{\text{proxy},(1)}$ (step 1). 
${D}^{\text{proxy},(1)}$ has no overlap with $D^{target}$ in terms of data samples and origin, and its distribution is different from $D^{target}$.
The inspector then uses the trained shadow model $f^\text{sha}(\cdot)$, with  ${D}^{\text{proxy},(2)}$ to generate a model access dataset, which we name reference set ${R}^\text{train}$ (step 2).  
${D}^{\text{proxy},(2)}$ contains some origins in ${D}^{\text{proxy},(1)}$ and some not, while ${D}^{\text{proxy},(1)} \cap {D}^{\text{proxy},(2)}=\varnothing$.
For whose origins are in ${D}^{\text{proxy},(1)}$, their origin membership labels in ${R}^\text{train}$ are positive; and for whose origins are not, their labels are negative. 
With featurized ${R}^\text{train}$ (step 3), the inspector then trains a meta classifier $g(\cdot)$, whose architecture is normally different from $f$.
$g$ is to learn the mapping between featurized reference data and origin memberships. 
This meta classifier $g(\cdot)$ is then exploited to infer the origin membership with the reference set $R^\text{test}$ generated from test dataset $D^\text{test}$ with the target model $f^\text{tar}(\cdot)$ (steps 5, 6 and 7).
The meta classifier $g$ can directly deliver the membership inspection (step 8), and the diversity and fairness metrics can be further analysed via Definition~\ref{def:diverse-inspect} and~\ref{def:fairness-inspect} (step 9) accordingly.

\subsubsection{Shadow Training}\label{sec:shadow-training}

In this work, shadow training (step 1,2,4) is used to train a shadow model $f^{sha}$, which has the same architecture as the target model $f$, and then analyse the behavior of $f^{sha}$ to get the mapping $g$ from access data of $f$ to origin membership in the inaccessible $D^{target}$.

The feasibility of this approach lies in that the training data of $f^{sha}$ is fully accessible.
We also assume the DNN architecture can be known in advance, as many cases mentioned in Section~\ref{sec:intro} currently work.
 Another assumption is that DNN with the same architecture have similar access output patterns regarding origin membership, which is similar with assumptions in~\cite{MIA, PIA,Miao2021TheAA,Song2019AuditingDP}.
Thus we can analyze the mapping between access output of $f^{sha}$ and the origin membership in $D^{proxy, (2)}$, and then transfer this mapping to $f$, predict origin membership in $D^{target}$ with the access output of $f$. For a set of auxiliary data $D^{aux}_v=X^{aux}_v\times Y^{aux}_v$ of origin $v$ ($D^{test}$ and $D^{proxy,(1)}$ are similar), the reference input data $r_v$ is 
\begin{equation}\small\label{eq:r}
    r_v = 
    \begin{cases}
    \FuncSty{Feat}(f(X^{aux}_v), D^{aux}_{v}), & X^{aux}_v \subset X^{test} \\
    \FuncSty{Feat}(f^{sha}(X^{aux}_v), D^{aux}_{v}), & X^{aux}_v \subset X^{proxy,(2)}
    \end{cases}
\end{equation}
\begin{equation}\small\label{eq:r-label}
    m_v = 
    \begin{cases}
    1, & v\in  V^{target} \vee v\in  V^{proxy,(1)}  \\
    0, & v\notin  V^{target} \wedge v \notin   V^{proxy,(1)}
    \end{cases}
\end{equation}

The reference data set is ${R}=\{(r_v, m_v)\}$, and \FuncSty{Feat} is origin-level Featurization function that will be introduced in Section~\ref{sec:mic}.

Given the two reference sets ${R}^\text{train}$ and ${R}^\text{test}$, inspectors can train and make origin inferences with a meta classifier $g$. 
Here the inspectors are encouraged to choose lighter-weight supervised learning models as the meta classifier $g$ than $f$. This is because the number of origins is typically much smaller than the number of data samples in the dataset, yielding a smaller reference set ${R}$ than with $D$.

\subsubsection{Featurization on Origin Level}\label{sec:mic}

Origin-level featurization enables the high-dimensional model access output to facilitate shadow training in Section~\ref{sec:shadow-training}.
As represented in Equation~\eqref{eq:r}, $f(D^{aux}_{v})$ is often high-dimensional. 
Furthermore, high-dimensional training data also introduces significant overhead when training $g$ and causes overfitting.

To overcome this problem, we exploited the statistical methods in to handle high-dimensional $f( D^{aux}_{v})$.
We employed the following three methods:
\begin{enumerate}
\item Concatenation: concatenate all elements in $f(\cdot, \{d^{aux}_{v}\})|d^{aux}_{v}\in D^{aux}_{v}\}$;
\item Statistics Tuple: include maximum, minimum, mean, $20^{th}$, $25^{th}$, $40^{th}$, $50^{th}$, $60^{th}$, $75^{th}$, $80^{th}$ percentile, variance and standard deviation values;
\item Histogram: probability density function of histogram for each $D^{aux}_{v}$ with a fixed number of bins.
\end{enumerate}

\begin{figure}[t]
    \centering
    \includegraphics[width=\linewidth]{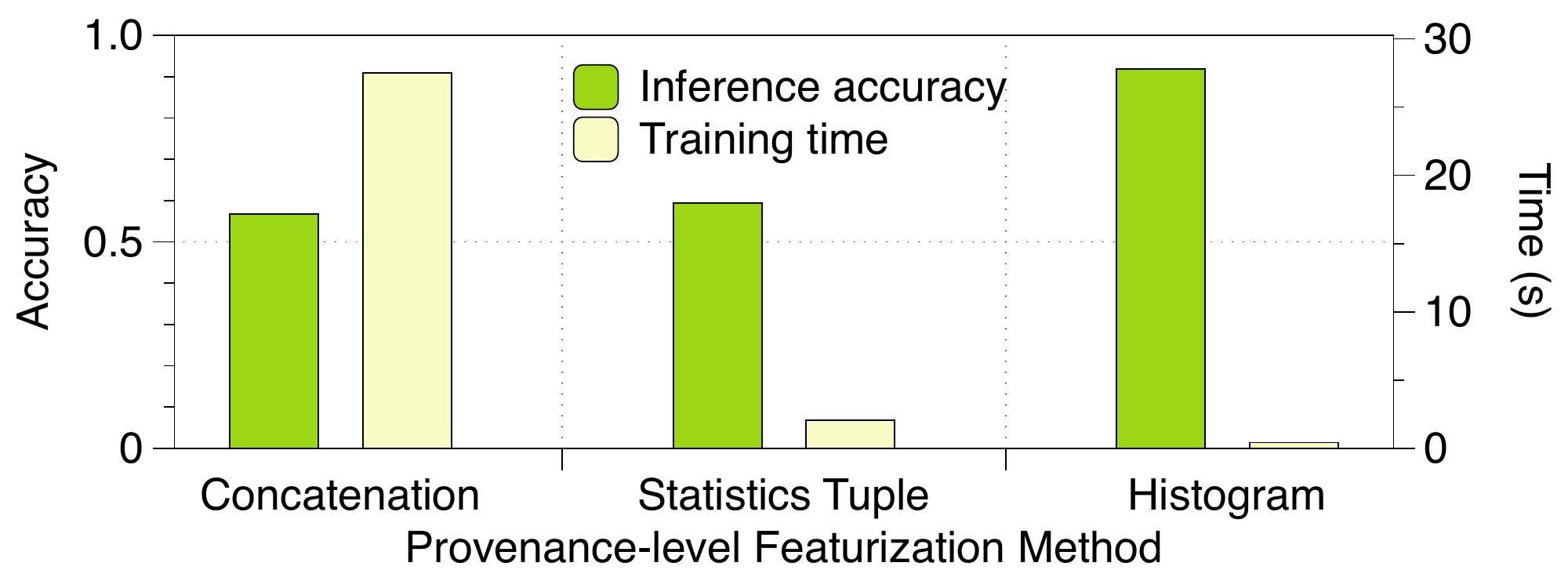}
    \caption{Our preliminary experiments demonstrates that Histogram significantly improves accuracy and reduces training time comparing to Concatenation and Statistics Tuple.}
    \label{fig:feat-yelp}
\end{figure}

Our preliminary experiment of these three methods illustrating in Figure~\ref{fig:feat-yelp} shows that the accuracy of the Histogram method outperforms the Concatenation and Statistics Tuple by 32.43\% while the training time reduces by 79.48\%.
\section{Case Study and Experiment Setup}\label{sec:exp}
To evaluate our framework in Section~\ref{sec:methodology} and\ref{sec:implementation}, we choose a real application as our case study and give the description of the experiment setup.
\subsection{Case Study: Movie Review Polarity Classification}
\label{subsec:dataset}
The application we would study is text polarity classification, which decides if the text is negative or positive. The dataset We use is IMDB Vision and NLP~\cite{xiaochenzhang_2022}, which is a movie review dataset (images and text) that contains 4, 067 movies and 12, 088 reviewers.

``Movie'' is selected the as the concerned data origin. The deep learning model we used is Small BERT~\cite{turc2019}, which is only a fewer encoder layers less than the typical BERT model\footnote{\url{https://tfhub.dev/google/collections/bert/1}}.  The sensitive attribute is gender. If $A=a$ in Definition~\ref{def:ori-behavior-bias} means ``the review involves gender-related vocabulary (e.g. female, girl, woman, etc).''
The underline aim of this setting is to investigate if the involvement of the protected group (in this case female) has impacts on the review polarity. Assuming $y=1$ means the review text is positive, if $b_v$ in Definition~\ref{def:ori-behavior-bias} is close to $1$, this means for the concerned data origin $v$, the text involving female tends to have a positive attitude while the male tends to have negative. If $b_v$ is close to $-1$, the female tends to have a negative while the male tends to have a positive. Only when $b_v$ is close to $0$, this data origin (in this case is movie) is fair and has no discrimination regarding gender. 

\subsection{Evaluation Metrics of Inspection Performance}\label{sec:metric}

According to Definition~\ref{def:diverse-inspect} and~\ref{def:fairness-inspect}, the Inspection of Diversity and Fairness is based on Data Origin Inspection. Therefore, a separate evaluation metric is used for Data Origin Inspection. On the other hand, since the core idea of the Inspection of Dispersion and Fairness are the sampling in quality assurance, the evaluation metric should reveal the accuracy of distribution estimation of the inaccessible training set.

\subsubsection{Accuracy and Precision for Data Origin Inspection}\label{subsec:eval-MI}

As mentioned in Section~\ref{subsec:small}, the Data Origin Inspection is used to check if the involved origins are consistent with the customers' personalised requirements. Thus the priorities of the evaluation are employed to check 1) if the prediction of the origin membership is correct, and 2) how confident when identifying an origin as the member. The former can be evaluated through accuracy, and the latter can be evaluated through precision. 

\subsubsection{Kolmogorov–Smirnov Test for Dispersion and Fairness Inspection}\label{subsec:eval-KS}

In Section~\ref{subsec:large} and~\ref{subsec:official} we propose two measurements in Definition~\ref{def:ios} and~\ref{def:ori-behavior-bias}, and their corresponding inspection - Diversity Inspection in Definition~\ref{def:diverse-inspect} and Fairness Inspection in Definition~\ref{def:fairness-inspect}.
As Diversity Inspection and Fairness Inspection are sampling-based schemes, the evaluation should be how well the known measurements of the predicted member origins represent the overall distribution of these two metrics in the target training set. 

Herein we adopted the Two-sample Kolmogorov–Smirnov (K-S) test to quantify the similarity between our estimated measurements and the distribution of the true measurements in the target training set. The K-S test has two outputs: K-S statistic $\Delta \mu$ and $p$-value, which both indicate if our estimations follow the same distribution as the ground truth in the target training set. The lower $\Delta \mu$, the higher $p$-value, the better our inspection.

\subsection{Baselines and Devices.}

We exploited random guess (RG) as our baseline. For the membership inspection (with measurement $H^{m}$ in Definition~\ref{def:ori}), the RG baseline randomly decides whether a data origin is the member data origin. For the data diversity and fairness inspection, the RG baseline estimates the data diversity (measured by $H^{m}$ in Definition~\ref{def:diverse-inspect}) and fairness (measured by $H^{e}$ in Definition~\ref{def:fairness-inspect}) based on the randomly predicted member data origin.

All experiments are conducted on a Ubuntu 20.04 server equipped with 1 24GB NVIDIA RTX A5000 GPU, 16 Intel Core i9-11900K @3.50GHz CPUs, and 128GB memory. Our implementation is based on TensorFlow 2.8.

\section{Results and Discussion}\label{sec:eval}
We next illustrate the inspection performance from the perspectives of modality, model layers, origin, fairness-related sensitive attributes and datasets. 

\begin{table*}[]\normalsize
\caption{Membership, diversity and fairness inspection on the review polarity classification. Layer types are Mask Embedding (ME), Token Embedding (TE), Position Embedding (PE), BERT encoder encoder (EE), BERT encoder pooled (EP), BERT encoder default (ED), BERT encoder sequence (ES),  Dropout (D) and Classifier (C). For the $H^{d}$ and $H^{e}$, the more the $p$-value is close to $1$, the more accurate the inspection is.}
\label{tab:imdb-res}
\begin{tabular}[h]{c|cc|c|ccc|cccc|c|c}
\toprule
\multirow{2}{*}{\textbf{Prov.\#}} & \multicolumn{2}{c|}{\multirow{2}{*}{\textbf{Metrics}}} & \multirow{2}{*}{\textbf{\begin{tabular}[c]{@{}c@{}}Random \\ Guess\end{tabular}}} & \multicolumn{3}{c|}{\textbf{0}} & \multicolumn{4}{c|}{\textbf{1}} & \textbf{2} & \textbf{3} \\ \cline{5-13}
& \multicolumn{2}{c|}{} &  & \multicolumn{1}{c|}{\textit{ME}} & \multicolumn{1}{c|}{\textit{TE}} & \textit{PE} & \multicolumn{1}{c|}{\textit{EE}} & \multicolumn{1}{c|}{\textit{EP}} & \multicolumn{1}{c|}{\textit{ED}} & \textit{ES} & \textit{D} & \textit{C} \\ \midrule
\multirow{6}{*}{102} & \multicolumn{1}{c|}{\multirow{2}{*}{$H^{m}$}} & accuracy & 0.5 & \multicolumn{1}{c|}{0.86} & \multicolumn{1}{c|}{\textbf{0.87}} & 0.86 & \multicolumn{1}{c|}{0.85} & \multicolumn{1}{c|}{0.80} & \multicolumn{1}{c|}{0.80} & 0.83 & 0.80 & 0.70 \\
& \multicolumn{1}{c|}{} & precision & 0.5 & \multicolumn{1}{c|}{0.74} & \multicolumn{1}{c|}{\textbf{1.00}} & \textbf{1.00} & \multicolumn{1}{c|}{\textbf{1.00}} & \multicolumn{1}{c|}{0.82} & \multicolumn{1}{c|}{0.82} & \textbf{1.00} & 0.82 & 0.93 \\ \cline{3-13}
& \multicolumn{1}{c|}{\multirow{2}{*}{$H^{d}$}} & $p$-value & 2.280e-10 & \multicolumn{1}{c|}{0.115} & \multicolumn{1}{c|}{0.153} & \textbf{0.161} & \multicolumn{1}{c|}{0.137} & \multicolumn{1}{c|}{0.088} & \multicolumn{1}{c|}{0.088} & 0.031 & 0.088 & 3.890e-6 \\
& \multicolumn{1}{c|}{} & $\mu$ & 0.408 & \multicolumn{1}{c|}{0.186} & \multicolumn{1}{c|}{\textbf{0.176}} & 0.175 & \multicolumn{1}{c|}{0.182} & \multicolumn{1}{c|}{0.197} & \multicolumn{1}{c|}{0.197} & 0.212 & 0.197 & 0.331 \\ \cline{3-13}
& \multicolumn{1}{c|}{\multirow{2}{*}{$H^{e}$}} & $p$-value & 0.659 & \multicolumn{1}{c|}{0.856} & \multicolumn{1}{c|}{0.772} & 0.874 & \multicolumn{1}{c|}{0.677} & \multicolumn{1}{c|}{0.848} & \multicolumn{1}{c|}{0.784} & 0.677 & 0.677 & \textbf{0.993} \\
& \multicolumn{1}{c|}{} & $\mu$ & 0.116 & \multicolumn{1}{c|}{0.093} & \multicolumn{1}{c|}{0.101} & 0.090 & \multicolumn{1}{c|}{0.112} & \multicolumn{1}{c|}{0.087} & \multicolumn{1}{c|}{0.101} & 0.112 & 0.112 & \textbf{0.054} \\ \midrule
\multirow{6}{*}{374} & \multicolumn{1}{c|}{\multirow{2}{*}{$H^{m}$}} & accuracy & 0.5 & \multicolumn{1}{c|}{0.82} & \multicolumn{1}{c|}{\textbf{0.83}} & 0.82 & \multicolumn{1}{c|}{0.81} & \multicolumn{1}{c|}{0.80} & \multicolumn{1}{c|}{0.80} & 0.82 & 0.80 & 0.78 \\
& \multicolumn{1}{c|}{} & precision & 0.5 & \multicolumn{1}{c|}{0.75} & \multicolumn{1}{c|}{0.70} & 0.72 & \multicolumn{1}{c|}{0.76} & \multicolumn{1}{c|}{0.67} & \multicolumn{1}{c|}{0.67} & 0.77 & 0.67 & \textbf{0.89} \\ \cline{3-13}
& \multicolumn{1}{c|}{\multirow{2}{*}{$H^{d}$}} & $p$-value & 1.020e-34 & \multicolumn{1}{c|}{3.500e-5} & \multicolumn{1}{c|}{6.340e-5} & 0.001 & \multicolumn{1}{c|}{\textbf{0.004}} & \multicolumn{1}{c|}{\textbf{0.004}} & \multicolumn{1}{c|}{0.001} & 1.680e-6 & \textbf{0.004} & 1.930e-9 \\
& \multicolumn{1}{c|}{} & $\mu$ & 0.399 & \multicolumn{1}{c|}{0.194} & \multicolumn{1}{c|}{0.192} & 0.168 & \multicolumn{1}{c|}{\textbf{0.163}} & \multicolumn{1}{c|}{\textbf{0.163}} & \multicolumn{1}{c|}{0.178} & 0.210 & \textbf{0.163} & 0.243 \\ \cline{3-13}
& \multicolumn{1}{c|}{\multirow{2}{*}{$H^{e}$}} & $p$-value & 0.000 & \multicolumn{1}{c|}{0.583} & \multicolumn{1}{c|}{0.532} & 0.903 & \multicolumn{1}{c|}{0.741} & \multicolumn{1}{c|}{\textbf{0.925}} & \multicolumn{1}{c|}{\textbf{0.925}} & 0.055 & \textbf{0.925} & 0.013 \\
& \multicolumn{1}{c|}{} & $\mu$ & 0.213 & \multicolumn{1}{c|}{0.068} & \multicolumn{1}{c|}{0.066} & \textbf{0.047} & \multicolumn{1}{c|}{0.059} & \multicolumn{1}{c|}{0.050} & \multicolumn{1}{c|}{0.050} & 0.106 & 0.050 & 0.119 \\ \bottomrule
\end{tabular}
\end{table*}
As introduced in Section~\ref{sec:metric}, the evaluation metrics for membership ($H^{m}$) are accuracy/precision, and that for diversity ($H^{d}$) and fairness ($H^{e}$) are the statics of Kolmogorov–Smirnov hypothesis test. The results of the text dataset are shown in Table~\ref{tab:imdb-res}. We implemented the dataset of two sizes of origins (102 and 374). Overall, the less involved the origins, the more accurate the inspection. $H^{d}$ tends to have higher accuracy in more shallow layers, but $H^{e}$ is more accurate in deeper layers. The following would analyse the three kinds of inspection.

\subsection{Data Origin Inspection $H^{m}$}
Where there are fewer data origins, the precision of the origin inference is higher than that of more origins. The accuracy decreases slightly when the number of origins increases, while the precision decreases more severely than the accuracy. When the involved origins are $102$, for some layers (e.g. Token Embedding, Position Embedding, BERT Encoder Encoder and BERT Encoder Sequence), the precision achieved $1.0$. This indicates that the fewer origins, the confidence of the membership origins are fairly high, nearly perfect. The highest inspection appears at the shallower layers, like the Token Embedding layer in the $0$th layer.

\subsection{Diversity Inspection $H^{d}$ \& Fairness Inspection $H^{e}$}
Both diversity inspection and fairness inspection have better performance around $0$th and $1$st layers, and the same as $H^{m}$, the fewer origins, the better performance.

Compared with fairness inspection, the diversity inspection has lower confidence about the distribution estimation, which is implied by the $p$-value in TABLE~\ref{tab:imdb-res}. Numerically, $H^{e}\in [-1,1]$ has a wider value range than $H^{d}\in [0,1]$, but the estimation errors of $H^{e}$ are more closed to $0$, as shown in Fig~\ref{fig:assumption}. This is consistent with what the $p$-value indicates in TABLE~\ref{tab:imdb-res}. For the same data origin, the fairness metric is more consistent than that of the data diversity metric, thus empirically Assumption~\ref{asm:fair-maintain} is more solid than Assumption~\ref{def:diverse-inspect}.

\section{Related Work}
\label{sec-related}

\subsection{Data Management in MLaaS}
In the field of data management and database, there is a concept similar to our proposed ``data origin'' - ``data provenance''~\cite{provenance}. 
However, ``data provenance'' usually contains the data processing or transforming process and involves the documentary of the metadata beforehand in terms of a specific type of data origins, which is not suitable for our case that the datasets might not be originally collected for the current customer's requests. 

\subsection{Training Set Information Inference} \label{sec:related-inf}
The studies of information inference of the training datasets usually fall into the scope of AI privacy. Here we introduce the most relevant two with our work - property inference and membership inference.

\subsubsection{Membership Inference}

Membership inference is to determine if a data item was used to train the target ML model~\cite{MIA}, and its prerequisite is that the data item is already given~\cite{ChoquetteChoo2020LabelOnlyMI,leino2019stolen, sablayrolles2019white}. 
For these given or explicit values in the datasets, there are variants of membership inference, like attribution inference and its extension correlation inference, which infer missing column values~\cite{fredrikson2014privacy} or the column correlation coefficients~\cite{crectu2021correlation} in tabular data, respectively. 

A relevant branch of membership inference to this work is a set-level extension of this technique, which checks user membership in text data~\cite{Song2019AuditingDP} and speech data~\cite{Miao2021TheAA}, where ``user" is a special case of data origin.

\subsubsection{Property Inference}
Property inference aims to infer properties about the training dataset, which may be unrelated to the model's original primary learning task.
There are centralized~\cite{ganju2018property, property-cnn} and decentralized~\cite{melis2019exploiting,orekondy2018gradient} scenarios. To implement such inference, there needs auxiliary data to learn a particular model output pattern of the existence of the concerned property, which methodologically differs from this work.

\subsection{Dataset Discrimination and ML Fairness}
ML fairness has been an increasing ethical topic in recent years since ML applications make essential decisions that may impact people's lives and careers~\cite{fairness-survey}. The definition of ``ML fairness'' is that there should be no bias or discrimination when the ML model makes decisions. The discrimination usually comes from a training set built up from discriminated data. These discriminated training data, in turn, lead to discriminated model decisions~\cite{fairness-survey}. As clarified in Definition~\ref{def:set-fairness} and Section~\ref{subsec:diversity-fairness}, we adopt group fairness, which refers to the protected group being treated similarly to the advantaged group or the other part of the population. We choose group fairness because it is the typical concern in the field of ML fairness and a suitable entry point for this work that first addresses this problem in the context of the private dataset of MLaaS. 
\section{Conclusion and Future Work}

This work is the first to investigate the problem of private dataset inspection, regarding dataset fairness and diversity. To this end, we first define the dataset diversity and fairness metrics and then build a dataset inspection framework. We then introduce a possible methodology with some empirical assumptions and afterwards propose a possible strategy that combines shadow training and multiple-instance learning. Our case study of review polarity classification has shown that our methodology is efficient.
For future work, there are two directions: 1) current inspection assumes the inspector has the ability to sample data origins randomly, it is worth considering if this assumption can be relaxed; 
2) In this work, we have discussed inter-origin diversity under inspection. A supplement to this would be a new metric quantifying intra-origin diversities. This metric can refine the origin distribution when customers only care about several particular origins.

\ifCLASSOPTIONcompsoc
  \section*{Acknowledgments}
\else
  \section*{Acknowledgment}
\fi

The authors would like to express their sincere gratitude to Prof. Julie A. McCann at Imperial College London. Her constructive suggestions greatly enhanced the quality of this work, also her efforts in gender equality encouraged the authors to put fairness in a critical role in this work.

\bibliographystyle{ieeetr}

\clearpage
% \appendix
% \input{section/appendix}

\end{document}